\documentclass[runningheads]{llncs}

\PassOptionsToPackage{table,xcdraw}{xcolor}

\usepackage{eccv}  % final/arXiv version

\usepackage{eccvabbrv}

\usepackage{graphicx}
\usepackage{booktabs}
\usepackage[accsupp]{axessibility}
\usepackage{makecell}
\usepackage{multirow}
\usepackage{tabularx}
\usepackage{algorithm}
\usepackage{algpseudocode}
\usepackage{siunitx}
\usepackage{adjustbox}
\usepackage{pifont}
\usepackage{threeparttable}
\usepackage{wrapfig}

\newcolumntype{Y}{>{\centering\arraybackslash}X}

\usepackage[breaklinks,colorlinks,citecolor=eccvblue]{hyperref}

\newcommand{\myruninhead}[1]{%
  \par\addvspace{\medskipamount}%
  \noindent\textbf{#1}}

\usepackage{orcidlink}

\begin{document}

% ---------------------------------------------------------------
\title{UECP: Uncertainty-Enhanced Collaborative Perception}

\titlerunning{UECP}

\author{Kang Yang\inst{1} \and
Tianci Bu\inst{2} \and
Peng Wang\inst{1} \and
Deying Li\inst{1} \and
Wen Jie\inst{3} \and
Yongcai Wang\inst{1}\thanks{Corresponding author.}}

\authorrunning{K.~Yang et al.}

\institute{Renmin University of China, Beijing, China\\
\email{\{yangkang1205,peng.wang,deyingli,ycw\}@ruc.edu.cn}
\and
College of Systems Engineering, National University of Defense Technology,
Changsha, China\\
\email{btc010001@gmail.com}
\and
Intelligent Shipping Center, China Waterborne Transport Research Institute,
Beijing, China\\
\email{wenjie@wti.ac.cn}}

\maketitle

\begin{abstract}
Collaborative perception serves as a pivotal solution to enhance the perception capability of individual agents in autonomous driving, where a core challenge lies in seeking reliable evidence to quantify and weight the contribution of each participating agent.
Existing methods typically rely on a confidence map (co-trained with the detection head), which is, however, inherently correlated with the detection results and thus fails to provide unbiased physical evidence. Furthermore, how to deeply integrate evidence into the cooperative fusion process remains an open question.
To address these issues, this paper first proposes \emph{uncertainty map}, a physically grounded and unambiguous metric for evaluating perception quality. This map is directly supervised by real-time sensor signals (i.e., LiDAR point density), ensuring decoupling from detection noise and thereby providing physical scenario-aware evidence for weighting agent contribution.
Based on this map, we develop the Uncertainty-Enhanced Collaborative Perception (UECP) framework, centered on the Uncertainty-Aware Pyramid Fusion (UAPF) module. UAPF uses a coarse-to-fine strategy, with two key components: Uncertainty-Weighted Downsampling (UWD) for high-fidelity feature preservation, and Uncertainty-Guided Residual Fusion (UGRF) to reinforce ego features, suppressing noise and ensuring robust fusion.
Extensive experiments on real-world datasets show UECP outperforms SOTA methods in effectiveness and robustness by embedding the uncertainty map into fusion. Code will be publicly available.
\end{abstract}

\begin{figure}[t]
    \centering
    \includegraphics[width=\linewidth]{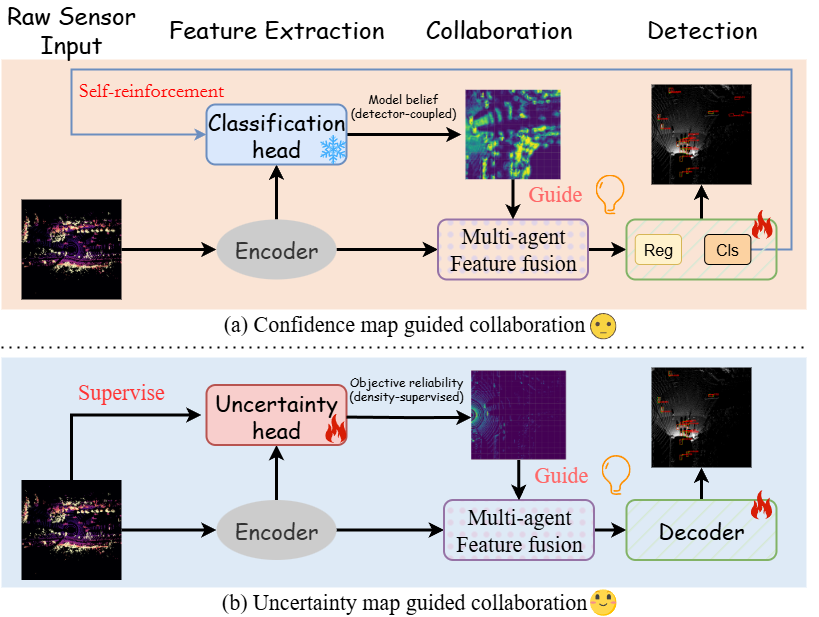}
    \caption{The difference between traditional confidence map and the proposed uncertainty map. The confidence map is co-learned with the classification head, while the uncertainty map is supervised by the LiDAR point density.}
    \label{intro_compare}
\end{figure}

\section{Introduction}
Collaborative perception is critical for autonomous driving, as it enables the ego agent to perceive non-line-of-sight hazards and thereby enhance driving safety \cite{cooperative-development1,cooperative-development2,where2comm,cooperative-development3,cooperative-development4,dair-v2x,OPV2V,v2v4real,v2x-seq}. A core challenge in collaborative perception lies in designing effective multi-agent information fusion strategies. Existing fusion approaches are generally categorized into early fusion \cite{Cooper1, disconet}, intermediate fusion \cite{OPV2V, where2comm, v2vnet, how2comm, when2comm}, and late fusion \cite{ROCO}, among which intermediate fusion—conducting feature-level fusion via compressed feature communication—achieves an optimal balance between communication efficiency and detection performance.
Within intermediate fusion, Bird's-Eye-View (BEV) feature fusion has become the dominant and most promising paradigm \cite{CoBEVT, v2x-vit}. However, mainstream BEV fusion methods \cite{Fcooper,v2vnet, OPV2V} often suffer from significant collaborative noise, which stems from view discrepancies, spatial misalignment, and other practical challenges. Developing a fusion mechanism that effectively mitigates such noise is therefore essential for achieving robust and high-precision perception.

\begin{figure}[t]
    \centering
    \includegraphics[width=\linewidth]{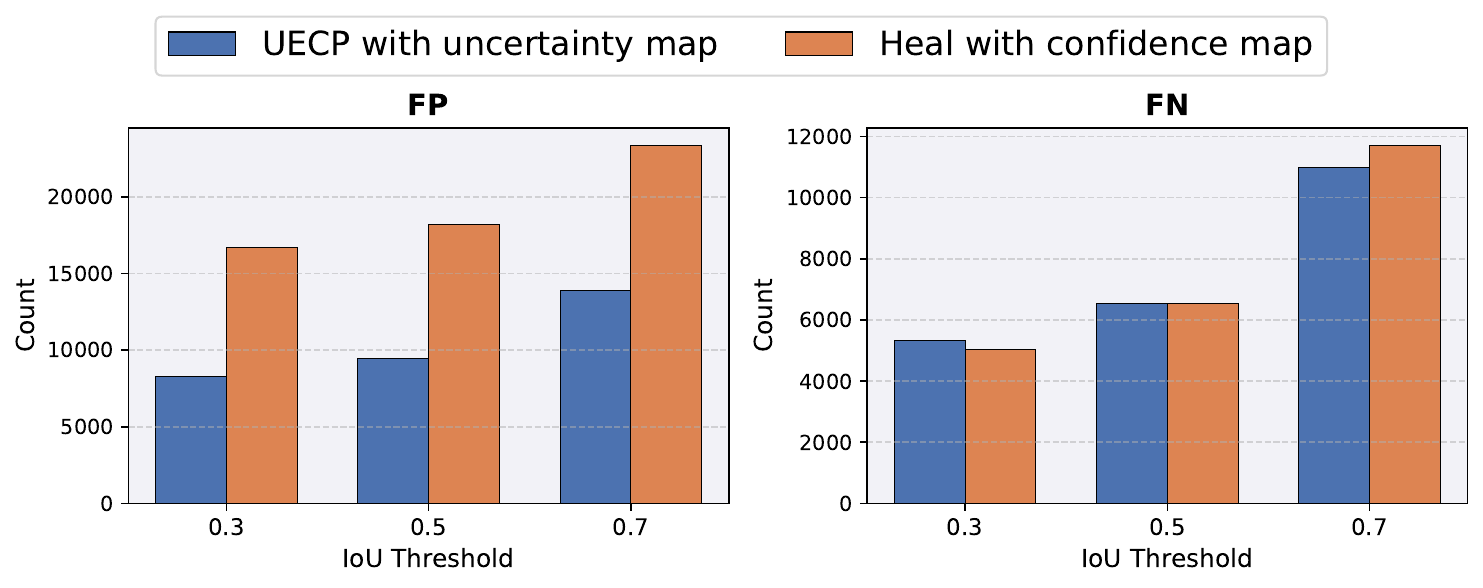}
    \caption{Comparison of false positives (FP) and false negatives (FN) between the proposed UECP and HEAL on the DAIR-V2X dataset under different IoU thresholds (0.3/0.5/0.7).}
    \label{Count}
\end{figure}

To address this noise issue, existing methods typically adopt confidence maps derived from the detector’s classification head to adaptively weight features during fusion \cite{HEAL,where2comm,TransIFF,ACCO,CodeFilling,cosdh}. Nevertheless, these confidence maps are inherently coupled with detection scores, failing to provide unbiased, physically grounded guidance for fusion. For example, they often struggle to suppress false positives, as demonstrated in \cref{Count}. Additionally, how to deeply integrate such guidance information into the fusion process remains an under-explored problem.

%To address these limitations, this paper firstly seeks physically grounded evidence. We propose to construct an uncertainty map derived from sensor signals (i.e., local LiDAR point cloud density) to provide objective evidence for collaborative feature fusion. \cref{intro_compare} illustrates the difference between confidence map and uncertainty map.
%The uncertainty map is learned from the point cloud density, which reflects the reliability of sensory observations. Crucially, diverse uncertainty sources---occlusion, truncation, and specular surfaces---all physically manifest as reduced point returns, making density a unified physical proxy rather than a narrow heuristic. Dense regions with abundant points indicate high observation confidence, whereas sparse regions correspond to higher uncertainty. In collaborative perception, high uncertainty does not imply ``cannot detect'' but rather signals ``need neighbor's information,'' enabling UGRF to down-weight unreliable ego regions and up-weight informative collaborators. The uncertainty map is decoupled from the detection head training and is more complementary to guide the fusion of the detection results.

To tackle these limitations, seeking physically grounded evidence is an important solution. For this purpose, \emph{uncertainty map} is proposed in this paper.  \cref{intro_compare} illustrates the fundamental difference between  the traditional confidence map and the proposed uncertainty map.  Uncertainty map is supervised by raw data feature, i.e., local LiDAR point cloud density. Because in LiDAR detections, diverse uncertainty sources (e.g., occlusion, truncation, specular surfaces) all physically manifest as reduced point returns, making point density a reasonal physical proxy of observation confidence. %Therefore, the learned uncertainty map reflects the reliability of sensory observations.  

With the uncertianty informaiton, in subsequent fusion, high uncertainty in collaborative perception does not mean “undetectable” but rather “in need of neighboring information”. This characteristic enables the fusion module to incooperate the uncertainty information into fusion process. More importantly, the uncertainty map is decoupled from the detection head training, making it more complementary to guiding fusion and improving detection performance.

Based on the uncertainty map, we further propose a novel multi-scale, uncertainty-involved fusion framework named Uncertainty-Enhanced Collaborative Perception (UECP). The core of UECP is an Uncertainty-Aware Pyramid Fusion (UAPF) module, which maximizes the utility of the uncertainty map through two key mechanisms.
First, we employ Uncertainty-Weighted Downsampling (UWD) to preserve high-quality information during the uncertainty-weighted pyramid construction, enhancing robustness to noise.
Second, and most critically, we introduce an Uncertainty-Guided Residual Fusion (UGRF) module at each pyramid scale. The UGRF generates a comprehensive fused feature representation for the ego-agent via an uncertainty-guided weighted summation, while its residual design stabilizes training by mitigating large variations in ego-centric features.

We perform extensive experiments to evaluate UECP on the real-world DAIR-V2X \cite{dair-v2x} and V2V4REAL \cite{v2v4real} datasets. Our results demonstrate that UECP consistently outperforms state-of-the-art methods, including HEAL~\cite{HEAL} and CoBEVT~\cite{CoBEVT}. Furthermore, our method retains its superior performance even under noisy conditions underscoring its inherent robustness.
In summary, our main contributions are three-fold:
\begin{itemize}
\item We propose uncertainty map, supervised by physical sensor signals, that acts as a robust and precise guidance signal for fusion in contrast to traditional confidence map.
\item We propose UECP, a comprehensive and novel fusion framework, which deeply involves uncertainty in the fusion process to overcome the limitations of confidence maps and the challenges posed by collaborative noise.
\item Extensive experiments on public real-world datasets validate that our proposed methods attain state-of-the-art (SOTA) performance across various precision metrics.
\end{itemize}

\section{Related Work}
\subsection{Cooperative Perception by Intermediate Fusion}
To overcome the perceptual limitations of individual agents, cooperative perception enables sensory information sharing across a network. The prevailing paradigm has gravitated toward intermediate fusion, which exchanges BEV feature maps to strike a balance between communication efficiency and fusion effectiveness. Intermediate fusion avoids the prohibitive bandwidth costs of early fusion (raw data) \cite{Cooper1,Cooper2} and outperforms the performance ceiling of late fusion. The core of this paradigm lies in the feature interaction mechanism. Initial works explored spatially-aware message passing using Graph Neural Networks (GNNs) \cite{v2vnet}, while AttnFuse \cite{OPV2V} first introduced attention to dynamically model inter-agent relationships. This line of research was significantly advanced by Transformer-based architectures like V2X-ViT \cite{v2x-vit}, which leverage powerful self-attention to learn global context from all collaborators. Concurrently, methods such as DiscoNet \cite{disconet} have explored knowledge distillation to enrich individual agents' features.

\subsection{Efficiency and Robustness in Fusion}
Despite their effectiveness, the dense feature exchange in these models poses substantial computational and bandwidth challenges. This has spurred a parallel research direction focused on sparse and efficient fusion. Where2comm \cite{where2comm} pioneered this area by learning communication priorities, selecting only the most salient feature regions. Subsequent works advanced this paradigm by focusing on instance-level features: TransIFF \cite{TransIFF} adopted a transformer-based approach, QUEST \cite{quest} proposed interpretable query cooperation, and ACCO \cite{ACCO} introduced sparse anchor collaboration---all to further reduce bandwidth overhead.
Beyond performance and efficiency, a key focus of recent research is ensuring the robustness and versatility required for real-world deployment. A large body of work has targeted enhancing robustness against systemic uncertainties inherent in cooperative systems, including specialized methods to mitigate pose error impacts \cite{localizationproblem, Coalign, ROCO, sparsealign, conformal_uncertainty} and compensate for communication latency \cite{latency1,latency2,Traf-align, CoBEVFlow}.
Building on these foundational robustness efforts, a further layer of complexity arises when collaborating agents are heterogeneous. Research in this space has explored interoperability across different sensor modalities: CoBEVT \cite{CoBEVT} and CoCa3D \cite{CoCa3D} extended the paradigm to camera-based systems, while V2VFormer++ \cite{v2vformer++}, CoCMT \cite{cocmt}, and BM2CP \cite{bm2cp} targeted multimodal suites. Concurrently, another research line, including HEAL \cite{HEAL}, HM-VIT \cite{HM-VIT}, and STAMP \cite{stamp}, focuses on ensuring feature compatibility when agents employ disparate internal models or data types. In this paper, we facilitate essential and supportive information exchange among agents guided by the uncertainty map.

\subsection{Uncertainty Modelling}
Uncertainty plays a critical role in safe autonomous driving. We distinguish between aleatoric uncertainty, stemming from sensor noise, environmental variability, and measurement errors, and epistemic uncertainty, which reflects model ignorance due to limited data or coverage gaps~\cite{uncertainty_01,uncertainty_two_para}.
To estimate epistemic uncertainty, methods such as Bayesian neural networks, Monte Carlo dropout, and deep ensembles approximate predictive distributions via sampling or ensembling~\cite{Bayesian_Neural_Networks,MC_Dropout,Deep_Ensembles}. In contrast, aleatoric uncertainty is often captured in a single forward pass through learned variance or estimated using evidential and conformal prediction techniques, which offer calibration during training or post hoc~\cite{EvidentialDeepLearning,ConformalPrediction}.
Although these approaches have proven effective in standard 3D detection and autonomous driving tasks~\cite{Uncertainty_DirectApplications,Uncertainty_DirectApplications2,Uncertainty_DirectApplications3,Uncertainty_DirectApplications4,Uncertainty_DirectApplications5}, their application to cooperative perception remains limited. Many existing methods rely on learned confidence scores~\cite{HEAL,where2comm,TransIFF,ACCO,CodeFilling,cosdh}, typically derived from model posteriors $p(\text{object}\mid\text{features})$.
However, confidence-based fusion has inherent limitations in collaborative settings, as model-derived confidence often fails to reflect the reliability of shared sensor evidence. This typically leads to two failure modes.
First, when observations are sparse, each agent's confidence may be low, and fusion discards weak yet consistent signals via thresholding---causing missed detections.
Second, fusion can amplify false positives, where a spurious high-confidence activation dominates message weighting.
These issues underscore a mismatch between confidence-based fusion and the demands of robust collaboration, motivating evidence-driven uncertainty representations that better reflect sensor quality.

\section{Method}

\subsection{Collaborative Perception With Uncertainty}
We consider a collaborative perception system consisting of $N$ agents, whose interactions are modeled by a communication graph $G$. Each agent $n$ is a vertex equipped with a unique sensor setup (e.g., LiDAR or camera) and captures local sensory input $\mathcal{X}_n$. While any agent can function as both an information sender and receiver, we focus on a common setting where a single agent $i$ (the ``ego-agent") acts as the receiver, and all other agents $j$ serve as senders. The goal is to enhance the ego-agent's 3D object detection performance by aggregating information from its collaborators. We denote $\mathcal{O}_i$ the ground-truth objects surrounding agent $i$. Our objective is to learn a collaborative perception model $\Phi_\theta$ based on $\{\mathcal{X}_i, \mathcal{U}_i, \{\mathcal{M}_{j\to i}\}_{j\in\mathcal{N}(i)} \}$ to maximize the detection gain $g(\text{prediction}, \text{groundtruth})$ for the ego-agent. The optimization problem with consideration of the observation uncertainty is formulated as:
\begin{equation}
    \max_{\theta} \quad g\left(\Phi_\theta\left(\mathcal{X}_i,\mathcal{U}_i, \{\mathcal{M}_{j\to i}\}_{j\in\mathcal{N}(i)} \right),\mathcal{O}_i\right),
\end{equation}
where $\mathcal{U}_i$ is the uncertainty clue of $\mathcal{X}_i$; $\mathcal{M}_{j\to i} = \mathcal{P}(\mathcal{X}_j, \mathcal{U}_j)$ denotes the message transmitted from agent $j$ to agent $i$ with the uncertainty information.
$\mathcal{N}(i)$ is the set of neighbors of agent $i$ in the graph $G$.
The key challenge is to optimize the perception model $\Phi_\theta$ and to properly design the uncertainty clues $\mathcal{U}_i$ to effectively fuse information from multiple agents.

\subsection{Uncertainty Map Generation}
To quantify the uncertainty of each agent's local perception, unlike existing methods that typically adopt a confidence map co-learned with the classification head, we propose a novel uncertainty map.
Specifically, the uncertainty map is learned via a dedicated uncertainty head, supervised directly by raw sensor data, to provide physical grounding for feature fusion.
We demonstrate that this physically grounded uncertainty information enhances model robustness and mitigates collaborative noise.
We next describe how the raw observation data is processed to build the supervision signal and how the uncertainty map is learned.

\myruninhead{BEV encoder.} For the $i$-th agent, given its input $\mathcal{X}_i$, we extract the BEV feature $\mathcal{F}_i = \Phi_{\text{enc}}(\mathcal{X}_i) \in \mathcal{R}^{H \times W \times C}$ using a BEV encoder $\Phi_{\text{enc}}$, where $H, W$ and $C$ denote the height, width, and channel dimensions, respectively.

\myruninhead{Uncertainty Model.}
The uncertainty head $\Phi_{\text{unc}}(\cdot)$ predicts the spatial distribution of sensing reliability from each agent's BEV feature.
Given the extracted BEV representation $\mathcal{F}_i$, it produces a dense uncertainty map:
\begin{equation}
    \mathcal{U}_i = \Phi_{\text{unc}}(\mathcal{F}_i) \in [0,1]^{H \times W},
\end{equation}
where $\mathcal{U}_i{[x,y]}$ indicates the uncertainty of grid $[x,y]$, with higher values indicating lower sensing reliability.
Importantly, the uncertainty map captures \emph{sensing reliability} rather than semantic objectness. Semantic relevance is handled by the detector's learned features; the uncertainty map only modulates how much to trust each agent's contribution during fusion. High-density background regions thus contribute reliable negative evidence (i.e., ``no object here''), which is equally valuable for suppressing false positives.

\myruninhead{Uncertainty Head Training.}
To train the uncertainty model $\Phi_{\text{unc}}(\cdot)$, raw LiDAR scans are projected onto a BEV grid with resolution $\Delta x = \Delta y = 0.4$m to compute a density map $\mathcal{D}_i$ by counting points per cell.
The map is normalized to $[0,1]$, and ground-truth uncertainty is defined as:
\begin{equation}
\mathcal{U}_i^{\text{gt}} = \mathbf{1} - \text{norm}(\mathcal{D}_i),
\end{equation}
where sparser regions indicate higher uncertainty.
The uncertainty $\mathcal{U}_i$ is trained toward $\mathcal{U}_i^{\text{gt}}$ using a hybrid loss that combines a focal-style regression term
to emphasize difficult regions and a gradient-consistency regularizer to preserve structural boundaries:
\begin{equation}
    \mathcal{L}_{\text{un}} = \mathcal{L}_{\text{focal}} + w_g \, \mathcal{L}_{\text{grad}},
\end{equation}
where $\mathcal{L}_{\text{grad}}$ is computed from fixed Sobel filters \cite{sobel}.
The full formulation and parameter settings are provided in the Appendix~\ref{loss_function}.
This learned uncertainty map subsequently serves as a guidance signal for collaborative fusion.

\begin{figure}[t]
    \centering
    \includegraphics[width=\linewidth]{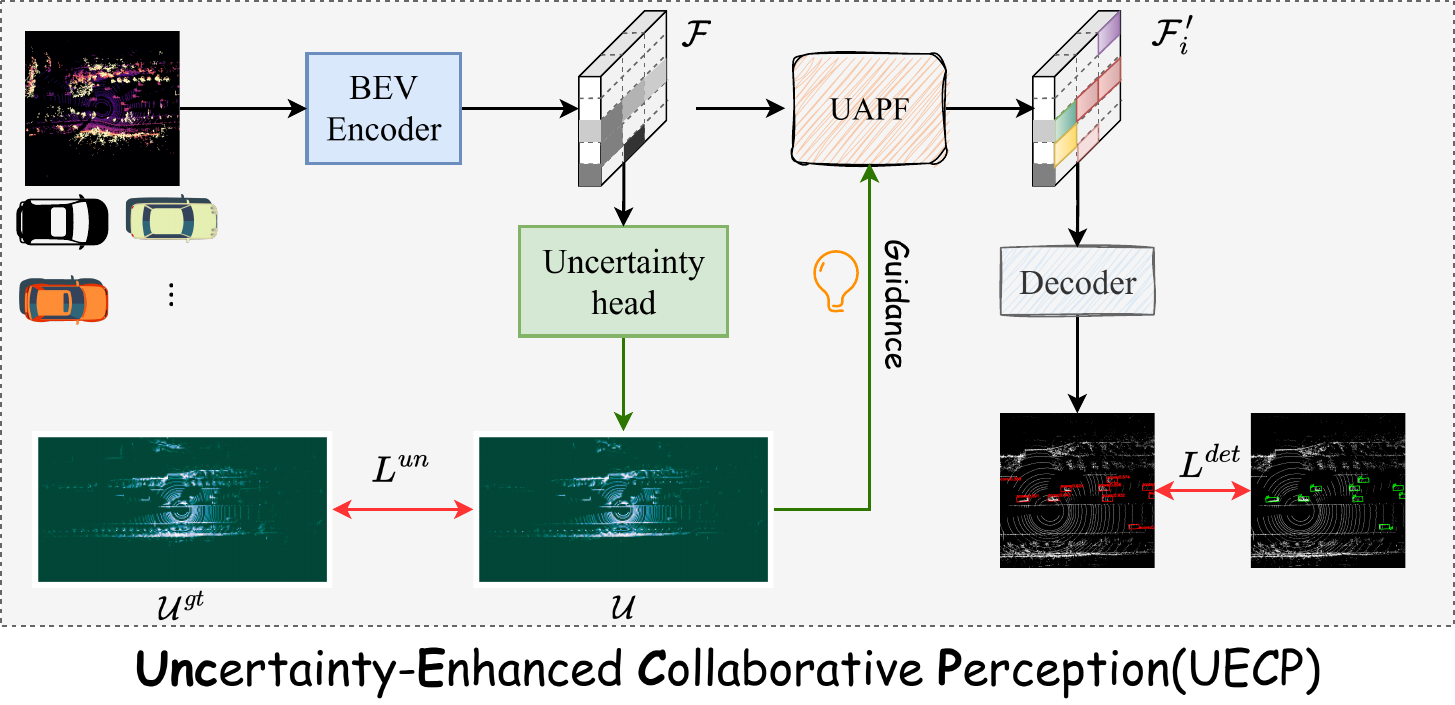}
    \caption{The overall pipeline of our proposed framework, UECP. The framework begins by processing raw sensor data from each agent independently through a shared BEV Encoder to extract initial features. In parallel, an uncertainty head predicts a physically-grounded uncertainty map for each agent. This map serves as the key input to our Uncertainty-Aware Pyramid Fusion (UAPF) module. The resulting deeply-fused feature map is finally passed to a detection head to generate the cooperative predictions.}
    \label{overview}
\end{figure}

\begin{figure}[t]
    \centering
    \includegraphics[width=\linewidth]{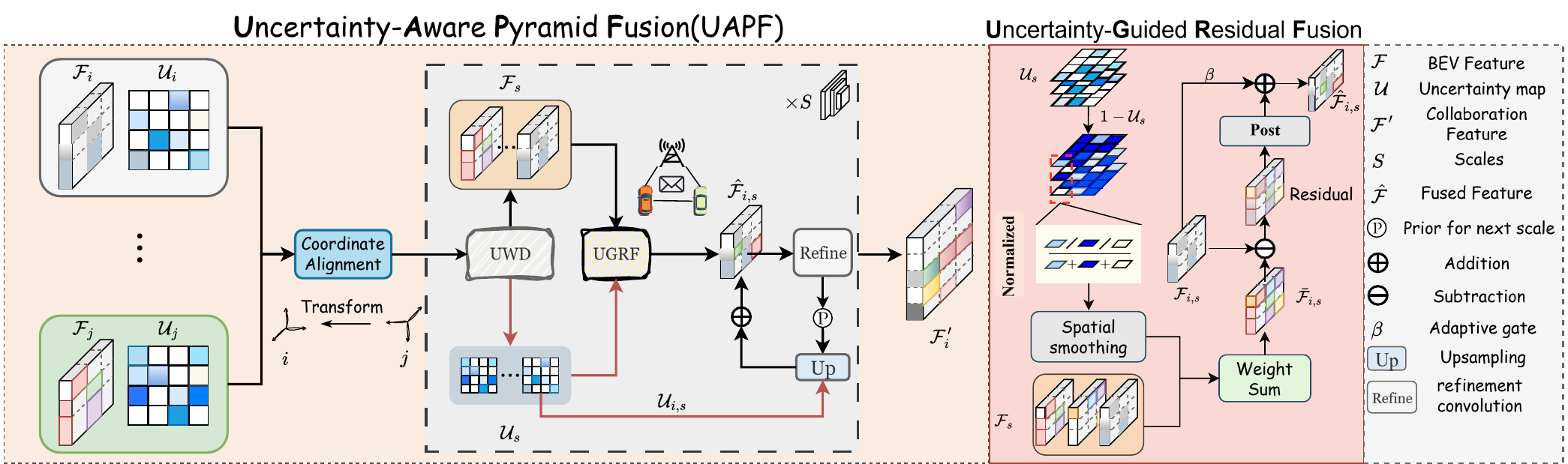}
    \caption{The overall pipeline of UAPF. Within the UAPF module, our Uncertainty-Weighted Downsampling (UWD) mechanism first constructs a high-fidelity feature pyramid. Then, at each scale, an Uncertainty-Guided Residual Fusion (UGRF) block robustly integrates collaborative information.}
    \label{Framework}
\end{figure}

\subsection{Uncertainty Enhanced Collaborative Perception}
Building upon the proposed uncertainty map, we further present the Uncertainty Enhanced Collaborative Perception (UECP) framework, as depicted in \cref{overview}.
UECP first feeds raw sensor data $\mathcal{X}_i$ into a BEV Encoder to produce feature $\mathcal{F}_i$, and subsequently outputs the uncertainty map $\mathcal{U}_i$ via the dedicated uncertainty head.
Thereafter, $\mathcal{F}_i$, $\mathcal{U}_i$, alongside the features $\mathcal{F}_j$ and uncertainty maps $\mathcal{U}_j$ of neighboring agents $j \in \mathcal{N}_i$, where $\mathcal{N}_i$ denotes the neighborhood of agent $i$, are transmitted via suitable neighborhood communication protocols \cite{HEAL, where2comm, CoBEVT}, a standard operation in multi-agent collaborative perception to ensure efficient and reliable feature fusion.
These aggregated inputs are then fed into UECP's core component, namely the Uncertainty-Aware Pyramid Fusion (UAPF) module, for final processing.

\myruninhead{Uncertainty-Aware Pyramid Fusion (UAPF).}
UAPF performs multi-scale collaborative fusion to facilitate effective multi-agent interaction.
UAPF integrates two key submodules: (1) the Uncertainty-Weighted Downsampling (UWD) module to enable the model to capture rich collaborative features across scales, from coarse to fine; and (2) the Uncertainty-Guided Residual Fusion (UGRF) block.
UWD extracts high-quality features while enhancing robustness through uncertainty-guided downsampling. Meanwhile, UGRF refines the receiving agent's features via an uncertainty-aware enhancement process, incorporating an ego-centric residual structure.
A detailed architectural illustration of UAPF is provided in \cref{Framework}. The process within the UAPF module unfolds as follows.

As a prerequisite for fusion, all uncertainty maps and BEV features are first transformed into a unified ego-coordinate frame \cite{ROCO}. The UAPF module then takes the aligned uncertainty map $\mathcal{U}_i$, BEV feature $\mathcal{F}_i$, and $\mathcal M_{j\to i}$ that encapsulates neighbor agents' feature and uncertainty $\{\mathcal{F}_j, \mathcal{U}_j\}, j\in \mathcal N(i)$ as input to produce the final collaborative feature $\mathcal{F}_i^{\prime}$.

The feature data and uncertainty map are firstly downsampled to multiple scales.
At each scale level $s$, we employ an Uncertainty-Weighted Downsampling (UWD) module, $\Phi_{\text{uwd}}$, to process the input feature map $\mathcal{F}$ and uncertainty map $\mathcal{U}$. This operation generates the downsampled outputs $(\mathcal{F}_s, \mathcal{U}_s) = \Phi_{\text{uwd}}(\mathcal{F}, \mathcal{U})$.
Subsequently, an Uncertainty-Guided Pyramid Fusion (UGRF) module processes these inputs to generate a fused feature map $\hat{\mathcal{F}}_{i,s}$ for the ego agent $i$. This fused feature is then post-processed by a convolutional refinement block to enhance its quality and informativeness.
To achieve multi-scale fusion, the refined feature $\hat{\mathcal{F}}_{i,s}$ is stored as a prior and upsampled to the next scale, where it is added element-wise to the newly fused feature. The final output $\mathcal{F}_i^{\prime}$ maintains the input BEV resolution.

\begin{figure}[t]
    \centering
    \includegraphics[width=\linewidth]{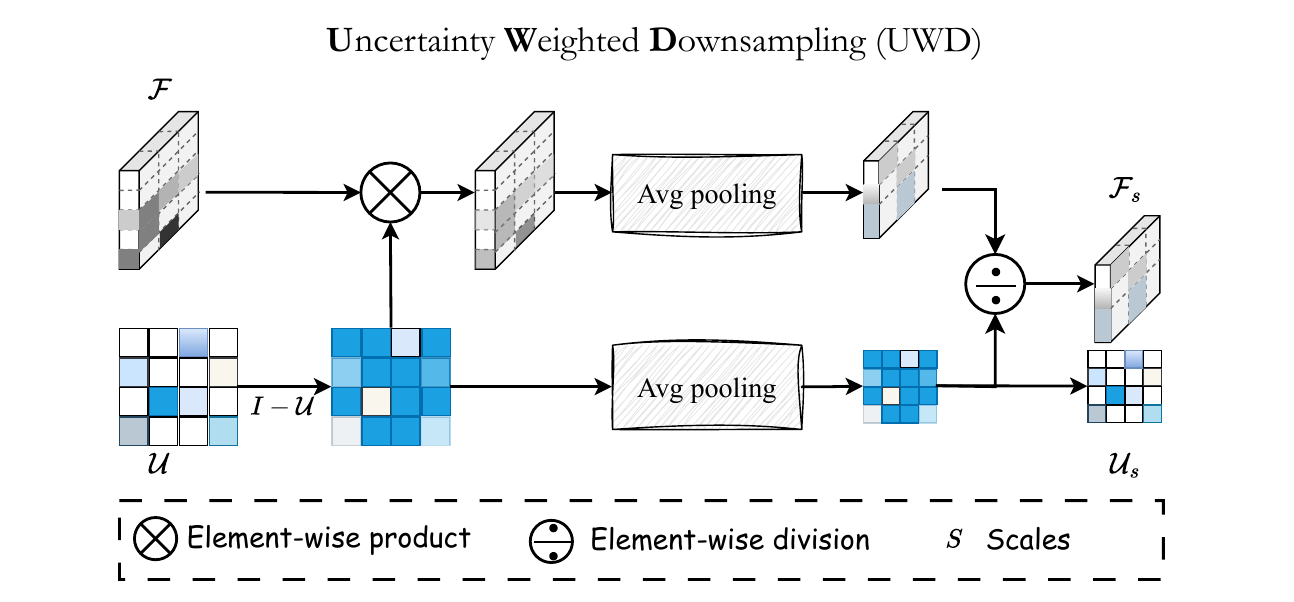}
    \caption{Uncertainty-Weighted Downsampling module.}
    \label{UWD}
\end{figure}

\myruninhead{Uncertainty-Weighted Downsampling (UWD).}
In a multi-scale feature pyramid, constructing lower-resolution representations that retain critical information from high-resolution inputs is essential for robust fusion.
Conventional downsampling methods such as max-pooling or average-pooling often suffer from noise sensitivity and information dilution \cite{downsample_01,downsample_02}. To overcome these limitations, we propose Uncertainty-Weighted Downsampling (UWD), which adaptively weighs feature contributions based on their reliability to preserve informative regions during the downsampling process, as illustrated in \cref{UWD}.
UWD takes a feature map $\mathcal{F}$ and an uncertainty map $\mathcal{U}$ as input and produces their downsampled counterparts $\mathcal{F}_s$ and $\mathcal{U}_s$. It is implemented as the ratio of two average-pooling operations, enabling efficient, uncertainty-aware aggregation with minimal overhead.
Given an uncertainty map $\mathcal{U}$, the downsampled feature is computed as:
\begin{equation}
\begin{split}
\mathcal{F}_s
&= \frac{\mathrm{AvgPool}_{s}(\mathcal{F} \odot (1-\mathcal{U}))}
         {\mathrm{AvgPool}_{s}(1-\mathcal{U})}, \\
&\text{i.e., for each location } (j,k): \\
(\mathcal{F}_s)_{jk}
&= \frac{\frac{1}{|\mathcal{W}_{jk}|}\sum_{p \in \mathcal{W}_{jk}} (1-\mathcal{U}_{(p)}) \cdot \mathcal{F}_{(p)}}
         {\frac{1}{|\mathcal{W}_{jk}|}\sum_{p \in \mathcal{W}_{jk}} (1-\mathcal{U}_{(p)})}.
\end{split}
\end{equation}
where $\mathrm{AvgPool}_s(\cdot)$ denotes average pooling with a given scale and stride,
$\mathcal{W}_{jk}$ represents the pooling window in the original feature map corresponding to location $(j,k)$ in the downsampled feature $\mathcal{F}_s$,
and $\odot$ is element-wise multiplication. The normalization factor cancels out, reducing the expression to a simple weighted average. This design leverages standard pooling operations to enable reliable, quality-aware downsampling with minimal modification.

\myruninhead{Uncertainty-Guided Residual Fusion (UGRF).}
To minimize collaborative fusion noise and ensure stable, consistent feature integration, we design the Uncertainty-Guided Residual Fusion (UGRF) module, a simple yet effective approach centered on residual fusion. For simplicity, the scale index is omitted.
Specifically, UGRF first generates a fusion weight $\lambda_n$ for each agent $n$, guided by its uncertainty map $\mathcal{U}_n$.
The fusion weight is then computed as
\begin{equation}
\label{lambda}
\lambda_{n} =
\frac{(\epsilon + \gamma \cdot (1-\mathcal{U}_n)) \odot v_n}
{\sum_{n=1}^{N}(\epsilon + \gamma \cdot (1-\mathcal{U}_n)) \odot v_n},
\end{equation}
where $\gamma > 0$ is a learnable scalar (enforced positive via $\gamma=\mathrm{softplus}(\tilde{\gamma})$),
$v_n \in \{0,1\}^{H \times W}$ denotes the visibility mask after ego-frame alignment and FOV cropping,
and $\epsilon = 10^{-8}$ ensures numerical stability.
Finally, a $3\times3$ Gaussian blur \cite{gaussian_blur} is applied to $\lambda$ to promote spatial smoothness.

Using the adaptive fusion weights $\lambda_n$,
the ego agent's fused representation is obtained through a weighted summation operation followed by a residual update:
\begin{equation}
\hat{\mathcal{F}}_{i}
= \mathcal{F}_{i}
+ \beta \cdot
\phi_\text{post}\!\left((
\underbrace{\sum_{n=1}^{N}\lambda_n \odot \mathcal{F}_{n})
- \mathcal{F}_{i}}_\text{Residual}
\right),
\end{equation}
where $\mathcal{F}_{i}$ denotes the ego agent's own feature,
and $\mathcal{F}_{n}$ represents the spatially aligned feature from the $n$-th collaborator.
$\phi_\text{post}$ is a shallow convolutional refinement network that enhances the collaborative residual,
and $\beta \in (0,1)$ is a learnable gate controlling the strength of the update.
This residual formulation preserves the ego feature as the primary representation,
while the aggregated information from collaborators acts as a controlled refinement signal.
Consequently, UGRF mitigates feature contamination from unreliable collaborators,
stabilizes training, and ensures a more robust and reliable fusion outcome.

\subsection{Loss function}
Finally, the training loss of the model is simply the sum of the regression loss $L_\text{reg}$, classification loss $L_\text{cls}$, direction loss $L_\text{dir}$ and uncertainty map loss $L_\text{un}$:
\begin{equation}
    L_\text{total} = \lambda_\text{reg} L_\text{reg} + \lambda_\text{cls} L_\text{cls} + \lambda_\text{dir} L_\text{dir} + \lambda_\text{un} L_\text{un},
\end{equation}
where the $\lambda$ are the weighting factors of the different losses used in the optimization process. We provide detailed illustrations for the loss function in Appendix~\ref{loss_function}.

\section{Experiments}
Training details, dataset descriptions, and benchmark configurations are provided in the supplementary material (Appendix~\ref{Exp}).

\subsection{Effectiveness of the uncertainty map}

\begin{table}[ht]
\centering
\caption{Replacing confidence map with uncertainty map consistently improves AP.
$\Delta$ = Unc $-$ Conf (pp).}
\label{exp_in_heal and where2comm}
\begin{adjustbox}{max width=\linewidth}
\begin{tabular}{lccccccccc c}
\toprule
\textbf{Method} &
\multicolumn{3}{c}{\textbf{AP@30}} & \multicolumn{3}{c}{\textbf{AP@50}} & \multicolumn{3}{c}{\textbf{AP@70}} & \textbf{Avg $\Delta$}\\
\cmidrule(lr){2-4}\cmidrule(lr){5-7}\cmidrule(lr){8-10}
 & Conf & Unc & $\Delta$ & Conf & Unc & $\Delta$ & Conf & Unc & $\Delta$ &  \\
\midrule
HEAL   & 79.11 & \textbf{80.60} & \color{green}+1.49 & 73.95 & \textbf{75.20} & \color{green}+1.25 & 55.56 & \textbf{58.93} & \color{green}+3.37 & \color{green}+2.03 \\
Where2comm   & 74.92 & \textbf{76.63} & \color{green}+1.71 & 67.90 & \textbf{71.48} & \color{green}+3.58 & 48.86 & \textbf{54.72} & \color{green}+5.86 & \color{green}+3.71 \\
\bottomrule
\end{tabular}
\end{adjustbox}
\vspace{-6pt}
\end{table}
\begin{figure}[t]
    \centering
    \includegraphics[width=\linewidth]{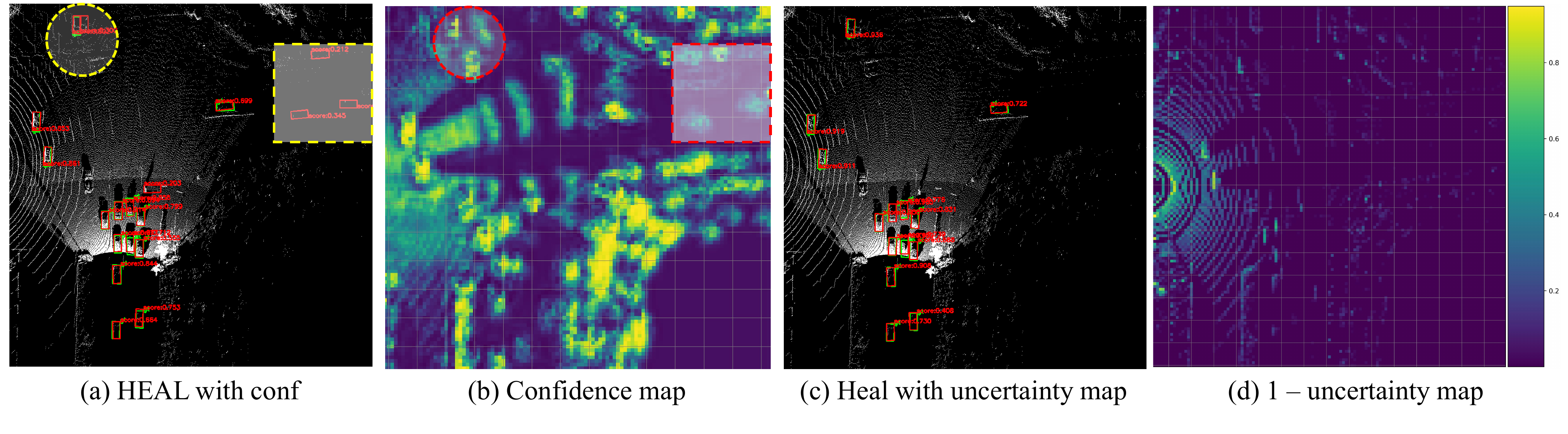}
    \caption{A comparative analysis of HEAL's performance guided by confidence versus that guided by uncertainty maps.}
    \label{vis_hm_vs_cm_in_heal}
\end{figure}

To validate the generality of our uncertainty map, we replace the confidence map with our uncertainty map in Where2comm and HEAL (\cref{exp_in_heal and where2comm}). The improvement is especially pronounced at IoU@0.7, indicating that the uncertainty map more effectively suppresses false positives. We attribute this to the confidence self-reinforcement problem: confidence maps are co-learned with the detector, so spurious high-confidence activations dominate fusion weighting and amplify false positives across agents. In contrast, the uncertainty map is decoupled from detection predictions and breaks this self-reinforcing cycle, providing an independent and physically grounded fusion signal. \cref{vis_hm_vs_cm_in_heal} further confirms this with qualitative comparisons, demonstrating consistent gains across different baseline architectures.

\begin{table}[!h]
\centering
\caption{Comparison of mainstream works on the V2V4REAL and DAIR-V2X dataset. The best performance is highlighted in \textbf{bold}, and the second-best performance is marked in {\color{blue}blue}.}
\small
\setlength{\tabcolsep}{4pt}
\begin{tabular}{c|ccc|ccc}
\Xhline{3\arrayrulewidth}
\cellcolor[HTML]{FFCCC9}                       & \multicolumn{3}{c|}{\cellcolor[HTML]{FFCCC9}V2V4REAL} & \multicolumn{3}{c}{\cellcolor[HTML]{FFCCC9}DAIR-V2X} \\ \cline{2-7}
\multirow{-2}{*}{\cellcolor[HTML]{FFCCC9}Method} & \multicolumn{1}{c|}{\cellcolor[HTML]{FFCCC9}AP@30} & \multicolumn{1}{c|}{\cellcolor[HTML]{FFCCC9}AP@50} & \multicolumn{1}{c|}{\cellcolor[HTML]{FFCCC9}AP@70} & \multicolumn{1}{c|}{\cellcolor[HTML]{FFCCC9}AP@30} & \multicolumn{1}{c|}{\cellcolor[HTML]{FFCCC9}AP@50} & \multicolumn{1}{c}{\cellcolor[HTML]{FFCCC9}AP@70} \\ \hline
No Coll                                         & 46.12 & 44.03 & 27.71 & 59.97  & 54.23 & 41.27\\
Fcooper                                         & 47.97 & 40.83 & 30.64 & 76.94 & 70.57 & 54.65 \\
Attn                                            & 48.18 & 40.06 & 28.75 & 69.83 & 64.06 & 48.98\\
V2VNet                                          & 45.12 & 43.24 & 31.75 & 69.25 & 65.52 & 47.84\\
V2X-VIT                                         & 40.11 & 37.18 & 24.53 & 77.28 & 72.47 & \color{blue}57.97 \\
CoBEVT                                          & \color{blue}63.13 & \color{blue}60.10 & 33.94 & 75.67 & 69.34 & 49.77 \\
Where2comm                                      & 49.61 & 45.10 & 34.45 & 74.92 & 67.90 & 48.86 \\
Who2comm                                        & 52.99 & 46.80 & 18.14 & 77.08 & 71.22 & 52.17\\
HEAL                                            & 51.57 & 48.70 & \color{blue}35.24 & 79.11 & 73.95 & 55.56 \\
CodeFilling                   & 60.15 & 59.78 & 35.07 & \color{blue}79.85 & \color{blue}75.18 & 57.26 \\
\hline
\rowcolor[HTML]{EFEFEF}
UECP &
\shortstack{\textbf{65.69}\\ {\color{green!60!black}(+2.56)}} &
\shortstack{\textbf{63.38}\\ {\color{green!60!black}(+3.28)}} &
\shortstack{\textbf{38.90}\\ {\color{green!60!black}(+3.66)}} &
\shortstack{\textbf{81.78}\\ {\color{green!60!black}(+1.93)}} &
\shortstack{\textbf{77.59}\\ {\color{green!60!black}(+2.41)}} &
\shortstack{\textbf{61.12}\\ {\color{green!60!black}(+3.15)}}  \\ \Xhline{3\arrayrulewidth}
\end{tabular}
\label{Performance}
\vspace{-6pt}
\end{table}

\subsection{Comparison with state-of-the-art}

\myruninhead{Detection Performance.} \cref{Performance} presents a comprehensive comparison of detection performance on the DAIR-V2X and V2V4REAL datasets. Our proposed method, UECP, achieves new state-of-the-art results across all metrics on both benchmarks. Notably, our approach demonstrates a substantial advantage at high-precision thresholds, achieving an AP@70 of 61.12\% on DAIR-V2X and 38.90\% on V2V4REAL.
On DAIR-V2X, UECP outperforms the second-best method by +1.93, +2.41, and +3.15 in AP@30, AP@50, and AP@70, respectively. The performance gains are even more pronounced on V2V4REAL, with improvements of +2.56, +3.28, and +3.66 across the same metrics.

\begin{figure}[t]
    \centering
    \includegraphics[width=\linewidth]{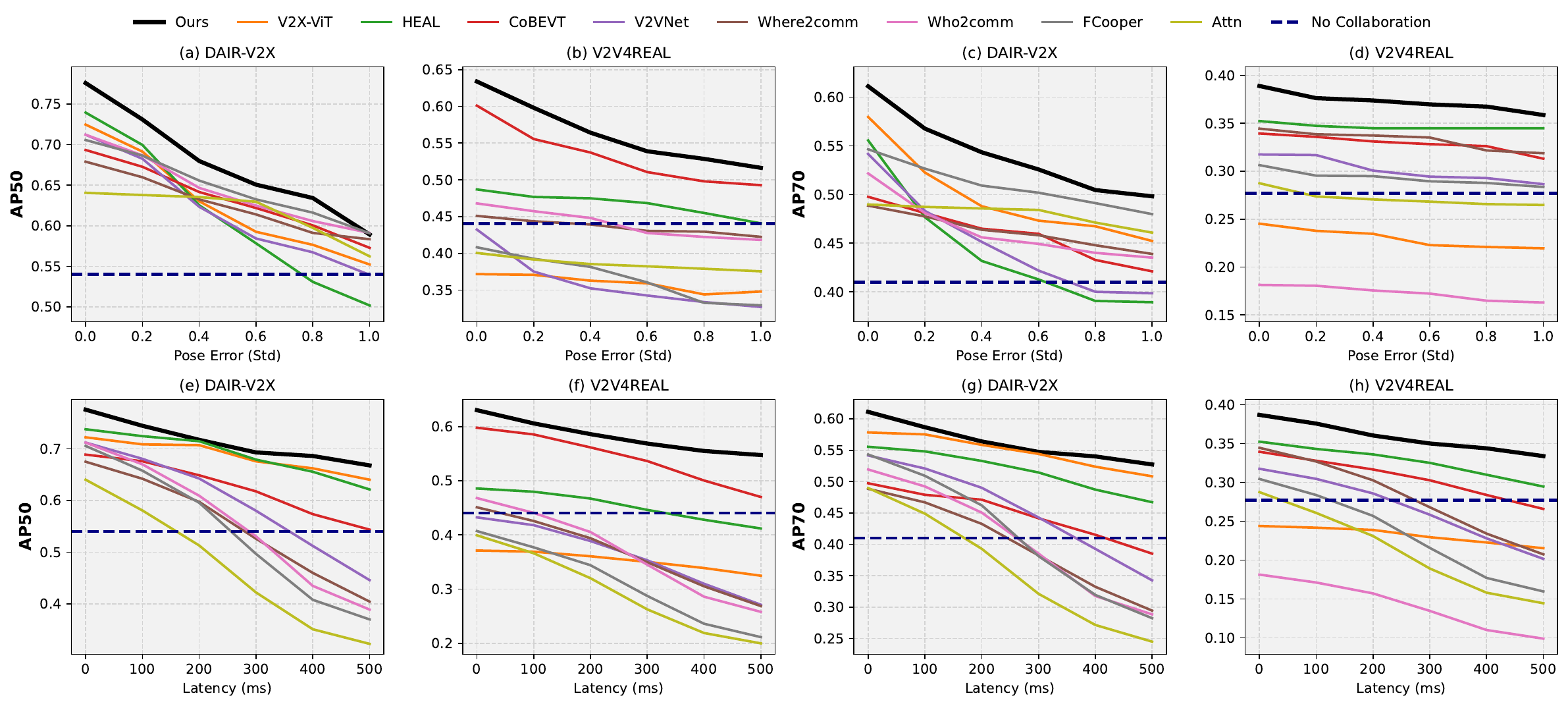}
    \caption{Robustness analysis under pose error and latency error on DAIR-V2X and V2V4REAL datasets.}
    \label{noise}
\vspace{-8pt}
\end{figure}

\myruninhead{Robustness.}
We evaluate robustness under pose noise ($\sigma$ from 0 to 1\,m, following CoAlign \cite{Coalign}) and communication latency (0--500\,ms, following SyncNet \cite{SyncNet}).
As shown in \cref{noise}, UECP maintains the highest accuracy under both disturbances. Communication-heavy methods degrade rapidly at high delay, whereas UECP's uncertainty-aware fusion preserves stable performance even under large temporal misalignment.

\subsection{Ablation studies}

\setlength{\intextsep}{0pt}%
\begin{wraptable}{r}{0.7\linewidth}
\centering
\caption{Ablation studies for proposed components.}
\label{ablation_components}
\small
\setlength{\tabcolsep}{3pt}
\begin{tabular}{cccc S[table-format=2.2] S[table-format=2.2] S[table-format=2.2]}
\toprule
\multicolumn{4}{c}{Components} & \multicolumn{3}{c}{Average Precision (AP)} \\
\cmidrule(lr){1-4} \cmidrule(lr){5-7}
 UM & MS & UWD & UGRF & {AP@0.3} & {AP@0.5} & {AP@0.7} \\
\midrule
            &             &             &              & 76.94 & 70.57 & 54.65 \\
\checkmark  &             &             &              & 78.52 & 72.39 & 56.92 \\
\checkmark  & \checkmark  &             &              & 79.98 & 75.73 & 59.74 \\
\checkmark  & \checkmark  & \checkmark  &              & 80.08 & 76.05 & 60.63 \\
\checkmark  & \checkmark  & \checkmark  & \checkmark   & \textbf{81.78} & \textbf{77.59} & \textbf{61.12} \\
\bottomrule
\end{tabular}
\end{wraptable}

In this subsection, we investigate the impact of each component in our method. All ablation results are reported on the DAIR-V2X validation set. We design two complementary experiments: \cref{ablation_components} isolates the effect of progressively adding each proposed component to a plain baseline (without UAPF), while \cref{ablation_three}(b) isolates the effect of the guidance-map type within the fixed UAPF architecture. Together, they confirm that both the uncertainty map and the UAPF architecture independently contribute to the overall gains.

\myruninhead{Effectiveness of our key components.} We conduct an ablation study to validate the effectiveness of our key components, with detailed results reported in \cref{ablation_components}. Our baseline, shown in the first row, emulates the F-Cooper setting by directly applying max fusion for collaboration. The subsequent rows demonstrate the progressive benefits of our contributions. First, incorporating our proposed uncertainty map yields a significant performance gain over the baseline. Building on this, the addition of the multi-scale (MS) pyramid architecture provides another substantial improvement, particularly at the stringent IoU@0.7 threshold. Finally, the results confirm that our two core modules, Uncertainty-Weighted Downsampling (UWD) and Uncertainty-Guided Residual Fusion (UGRF), each consistently and incrementally enhance the final detection performance.

\begin{table}[htbp]
\centering
\caption{Design choice ablation of pyramid scales, map types, and fusion operations in the fusion module.}
\label{ablation_three}
\vspace{-6pt}
\begin{adjustbox}{max width=\linewidth}
\begin{tabular}{ccc}

\begin{minipage}[t]{0.3\textwidth}
\centering
\subcaption{Effect of pyramid scales.}
\resizebox{\linewidth}{!}{
\begin{tabular}{c c c c}
\toprule
Scales  & AP@30 & AP@50 & AP@70 \\
\midrule
(1)     &  80.03  &  75.43  &  59.03  \\
(2,1)   &  81.00  &  76.12  &  59.79  \\
(4,2,1) &  \textbf{81.78}  &  \textbf{77.59}  &  \textbf{61.12}  \\
\bottomrule
\end{tabular}
}
\end{minipage} &

\begin{minipage}[t]{0.3\textwidth}
\centering
\subcaption{Effect of map type.}
\resizebox{\linewidth}{!}{
\begin{tabular}{c c c c}
\toprule
Map         & AP@30 & AP@50 & AP@70 \\
\midrule
No          &  77.89  &  72.63  &  55.91  \\
Confidence  &  78.68  &  73.60  &  56.47  \\
Uncertainty &  \textbf{81.78}  &  \textbf{77.59}  &  \textbf{61.12}  \\
\bottomrule
\end{tabular}
}
\end{minipage} &

\begin{minipage}[t]{0.3\textwidth}
\centering
\subcaption{Effect of fusion op.}
\resizebox{\linewidth}{!}{
\begin{tabular}{c c c c}
\toprule
Operation & AP@30 & AP@50 & AP@70 \\
\midrule
Max  & 80.93 & 76.01 & 59.86 \\
Mean & 80.33 & 75.13 & 57.66 \\
Sum  & \textbf{81.78} & \textbf{77.59} & \textbf{61.12} \\
\bottomrule
\end{tabular}
}
\end{minipage}

\end{tabular}
\end{adjustbox}
\vspace{-6pt}
\end{table}

\myruninhead{Fine-grained ablation.} We conduct a fine-grained ablation study to investigate the design effectiveness of our fusion module across three critical aspects, with detailed results presented in \cref{ablation_three}. As shown in \cref{ablation_three}(a), a hierarchical, multi-scale architecture (using scales 4, 2, and 1) achieves the best results, confirming the benefits of our approach. The results in \cref{ablation_three}(b) demonstrate that for the guidance signal, our proposed uncertainty map significantly outperforms both an un-guided baseline and a conventional confidence map, highlighting its superior ability to filter noisy features. We further compare the learned uncertainty map against using the density-based ground truth directly at inference in Appendix~\ref{learned_vs_density}; the learned map consistently outperforms density-direct, confirming that end-to-end training captures feature-level ambiguity beyond raw point counts. Finally, \cref{ablation_three}(c) analyzes the fusion operation, revealing that additive fusion (\texttt{sum}) yields the highest AP.

\begin{figure}[htbp]
    \centering
    \includegraphics[width=\linewidth]{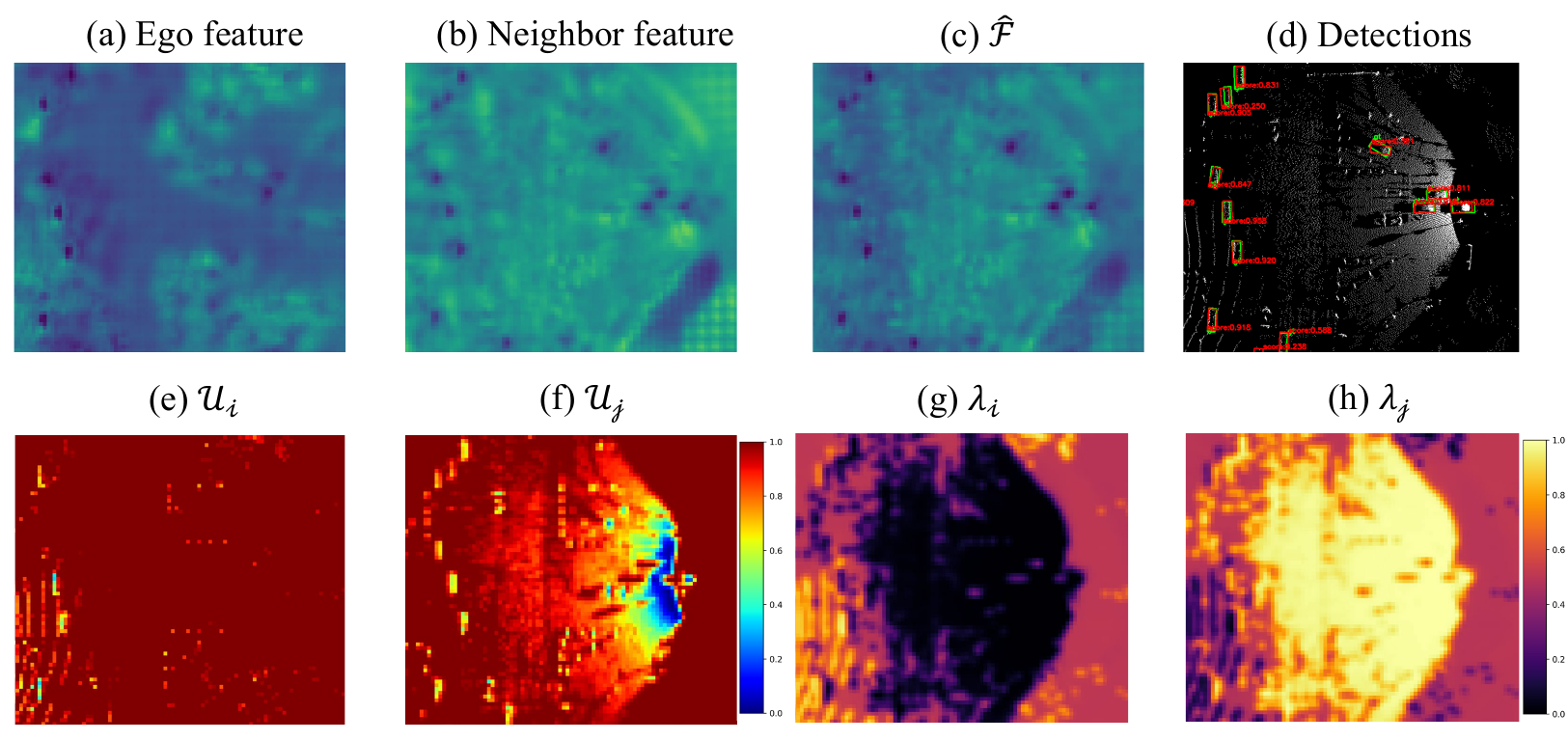}
    \caption{Visualization of collaboration in UECP. {\color{green}Green} and {\color{red}red} denote ground truth and detection, respectively.}
    \label{vis_analyse}
\end{figure}

\subsection{Qualitative evaluation}
Visualizations of detection results are provided in the supplementary material (Appendix~\ref{visualization of results}).

\myruninhead{Visualization of information selection and representation.} \cref{vis_analyse} illustrates the collaborative process of UECP. Subfigures (a)--(d) show the fused feature map $\hat{\mathcal{F}}$ and detection results in the ego coordinate system, confirming that neighbors provide complementary cues. Subfigures (e)--(f) display the uncertainty maps of the ego and neighbor, which capture object-level uncertainty and guide fusion. Subfigures (g)--(h) depict $\lambda$ derived from the uncertainty map (\cref{lambda}), highlighting inter-agent complementarity.

\section{Conclusion}
We propose UECP, a novel physical evidence-guided collaborative 3D detection system, which introduces three key components: UAPF, UWD, and UGRF.
A central contribution of UECP is the explicit generation and utilization of an uncertainty map, which provides physically grounded cues to guide collaboration.
Extensive experiments in real-world scenarios demonstrate that UECP not only achieves state-of-the-art perception performance but also remains robust under pose errors and communication latency.

% \myruninhead{Limitations.} We plan to explore whether the uncertainty map can be effective in scenarios where LiDAR quality is poor, as well as how it can be applied to multimodal settings. At the same time, we plan to adopt a more lightweight communication strategy.

% ---- Bibliography ----
\bibliographystyle{splncs04}
\bibliography{main}

\clearpage
% ---- Appendix ----
\appendix
\renewcommand{\theHsection}{A\arabic{section}}
\renewcommand{\theHsubsection}{A\arabic{section}.\arabic{subsection}}

\section{Method Details}

\subsection{Loss function}
\label{loss_function}
The training loss of the model is simply the sum of the regression loss $L_\text{reg}$, classification loss $L_\text{cls}$, direction loss $L_\text{dir}$ and uncertainty map loss $L_\text{un}$:
\begin{equation}
    L_\text{total} = \lambda_\text{reg} L_\text{reg} + \lambda_\text{cls} L_\text{cls} + \lambda_\text{dir} L_\text{dir} + \lambda_\text{un} L_\text{un},
\end{equation}
where the $\lambda$ are the weighting factors of the different losses used in the optimization process. In this paper, we set $\lambda_\text{reg} = 2$, $\lambda_\text{cls}=1$, $\lambda_\text{dir} = 0.2$ and $\lambda_\text{un} = 1$. We detail the loss function components as follows:
\begin{itemize}
\item $L_\text{reg}$: We use the L1 loss as the regression loss.
\item $L_\text{cls}$: To address the imbalance between foreground and background, we use the focal loss \cite{focal_loss} for classification.
\item $L_\text{dir}$: We formulate direction prediction as a classification task and apply the cross-entropy loss.
\item $L_\text{un}$: This loss targets two common issues in uncertainty-map regression: (i) value imbalance (many easy pixels vs.\ few hard ones) and (ii) over-smoothed details. It combines a continuous focal regression term with a gradient difference term computed via fixed Sobel filters \cite{sobel}:
\begin{equation}
\begin{aligned}
L
&= \alpha \cdot
   {\max\!\left(|e|^{\gamma+1}\right)}_{\text{continuous focal (regression)}} \\
&\quad + (1-\alpha)\cdot w_g \cdot
   \underbrace{\Bigl(\|\nabla_x \hat{y}-\nabla_x y\|_1+\|\nabla_y \hat{y}-\nabla_y y\|_1\Bigr)}_{\text{gradient difference}} .
\end{aligned}
\end{equation}
Where $e = \hat{y} - y$, $\hat{y}$ is the prediction, and $y$ is the ground truth. The focal exponent $\gamma$ down-weights easy pixels and emphasizes large errors; the gradient term preserves edges and fine structures. We set $\alpha = 0.5, \gamma=2.0$ and $w_g = 1.0$.
\end{itemize}

\section{Experiments}
\label{Exp}
\subsection{Setup}
\myruninhead{Dataset.} We evaluate our proposed method on two widely-used, real-world cooperative perception benchmarks: \textbf{DAIR-V2X} \cite{dair-v2x} and \textbf{V2V4REAL} \cite{v2v4real}.
The \textbf{DAIR-V2X} dataset is the first large-scale, multi-modal benchmark for Vehicle-Infrastructure Cooperative Autonomous Driving (VICAD), comprising over 71k LiDAR and camera frame pairs from real-world scenarios. The \textbf{V2V4REAL} dataset is specialized for Vehicle-to-Vehicle (V2V) cooperative detection in urban environments, spanning 410 km of driving routes. It contains 20k LiDAR frames, 40k RGB frames, and 240k annotated 3D bounding boxes across five categories, along with corresponding HDMaps.
For all experiments on these datasets, we set the perception range to $x \in [-102.4\text{m}, 102.4\text{m}]$ and $y \in [-51.2\text{m}, 51.2\text{m}]$, with a communication distance of 100m.

\myruninhead{Training details.} All models are trained within the same codebase for 60 epochs on four RTX 4090 GPUs, employing the one-cycle learning rate strategy \cite{one-cycle-lr} with AdamW \cite{adamaW} as the optimizer. The point clouds are segmented at a grid size of $0.4m \times 0.4m$. All detection models employ PointPillar \cite{PointPillars} as the backbone to extract 2D features from the point cloud.

\myruninhead{Baselines.} We compare against single-agent perception relying on the ego vehicle's data, alongside eight SOTA intermediate fusion models: Fcooper \cite{Fcooper}, Attn \cite{OPV2V}, V2VNet \cite{v2vnet}, V2X-VIT \cite{v2x-vit}, CoBEVT \cite{CoBEVT}, Where2comm \cite{where2comm}, Who2comm \cite{Who2comm}, CodeFilling \cite{CodeFilling}, and HEAL \cite{HEAL}.

\subsection{Learned Uncertainty Map vs.\ Density-Direct}
\label{learned_vs_density}

To validate the benefit of learning the uncertainty map end-to-end rather than using the raw point-density ground truth directly at inference, we compare four guidance-map variants within the same UAPF architecture on the DAIR-V2X validation set (\cref{tab:map_variant}). The learned uncertainty map $\hat{U}$ consistently outperforms the density-direct ground truth across all IoU thresholds, confirming that end-to-end training captures feature-level ambiguity beyond raw point counts. Both uncertainty-based variants substantially surpass the confidence map, which suffers from self-reinforcing false positives during fusion.

\begin{table}[h]
\centering
\caption{Comparison of guidance-map variants on DAIR-V2X (within the same UAPF architecture).}
\label{tab:map_variant}
\small
\setlength{\tabcolsep}{6pt}
\begin{tabular}{l c c c}
\toprule
Guidance Map & AP@0.3 & AP@0.5 & AP@0.7 \\
\midrule
No map          & 77.89 & 72.63 & 55.91 \\
Confidence      & 78.68 & 73.60 & 56.47 \\
Density-direct  & 81.73 & 77.28 & 60.77 \\
Learned $\hat{U}$ (Ours) & \textbf{81.78} & \textbf{77.59} & \textbf{61.12} \\
\bottomrule
\end{tabular}
\end{table}

\subsection{Model Efficiency Benchmarking}
\begin{figure}[t]
    \centering
    \includegraphics[width=\linewidth]{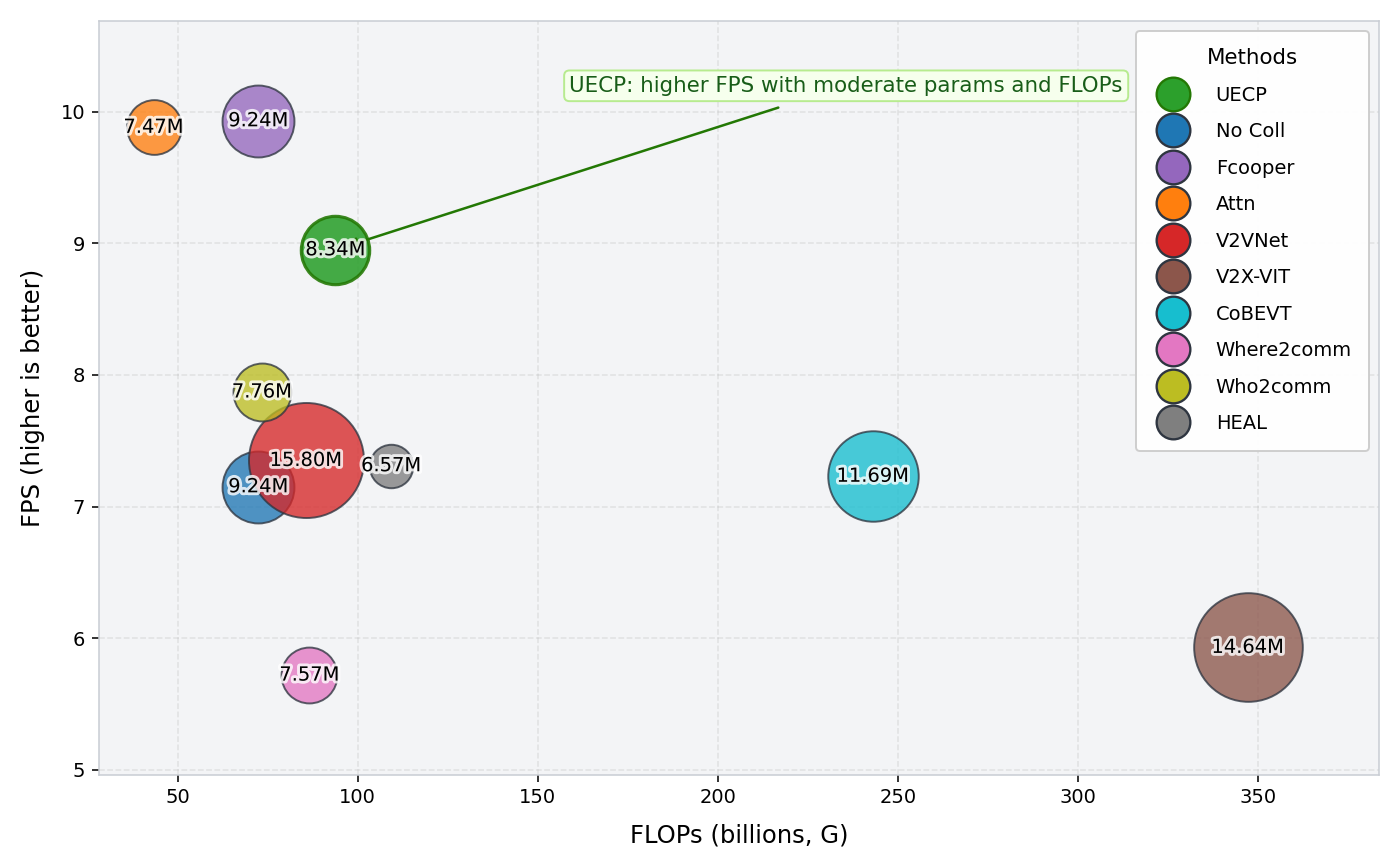}
    \caption{Comparison of FLOPs, FPS, and parameter size across collaborative perception methods. Each circle represents a different method, with its size proportional to the number of parameters (larger circles indicate more parameters). The x-axis denotes computational cost (FLOPs, in billions), and the y-axis shows inference speed (FPS, higher is better). UECP achieves a favorable balance, as highlighted in green.}
    \label{FPS}
\end{figure}

To evaluate the efficiency of various collaborative perception frameworks, we compare their computational complexity (FLOPs), model size (parameter count), and inference speed (FPS) in \cref{FPS}. As illustrated, traditional grid-based fusion methods such as V2VNet, CoBEVT, and V2X-ViT exhibit high computational costs and parameter counts, leading to reduced runtime efficiency. On the other hand, lightweight, communication-efficient approaches (e.g., Attn, F-Cooper) achieve faster inference but often sacrifice representational richness, resulting in suboptimal performance.
Our proposed UECP framework demonstrates an optimal trade-off among these factors. With only 8.34 M parameters and approximately 100 G FLOPs, UECP attains 9 FPS, surpassing most existing baselines in both computational efficiency and real-time performance.

\myruninhead{Communication overhead.} UECP transmits one additional uncertainty-map channel alongside the shared BEV features. At a typical feature dimension of $C{=}256$, this amounts to $1/256 \approx 0.39\%$ extra bandwidth, a negligible overhead. Moreover, the uncertainty map is orthogonal to communication-efficient frameworks such as Where2comm~\cite{where2comm} and Who2comm~\cite{Who2comm}, meaning it can be readily integrated without modifying their sparse communication strategies.

\subsection{Visualization of detection results}
\label{visualization of results}
\begin{figure}[t]
    \centering
    \includegraphics[width=\linewidth]{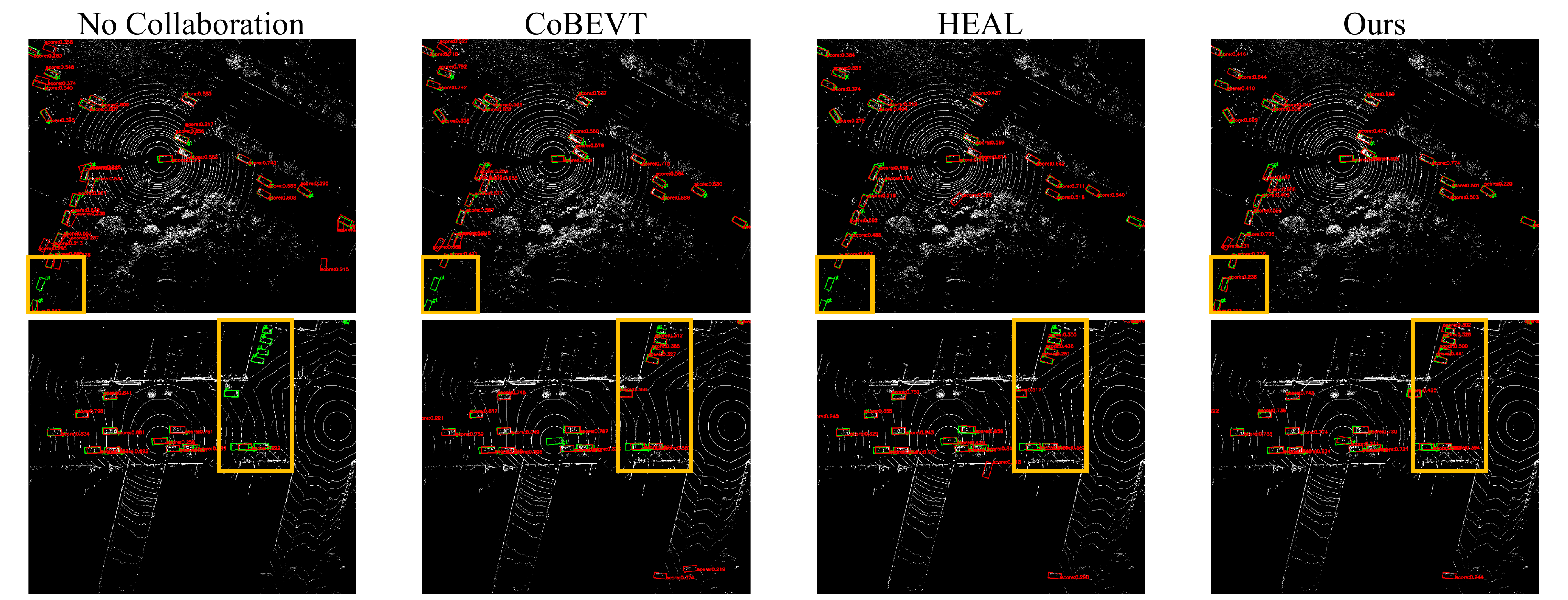}
    \caption{UECP achieves more accurate detections and less false positive. The first row corresponds to samples from the DAIR-V2X dataset, while the second row shows samples from the V2V4REAL dataset. {\color{green}Green} and {\color{red}red} boxes denote ground-truth and detection, respectively.}
    \label{result_vis}
\end{figure}

\cref{result_vis} provides a qualitative comparison with previous state-of-the-art (SOTA) methods. The visualization illustrates how our method, UECP, effectively avoids redundant information from collaborators and instead selectively transmits high-quality information. This leads to a superior detection output with fewer false positives and a higher rate of true positives compared to other approaches.

\end{document}